\pdfoutput=1

\documentclass[11pt]{article}

\usepackage[]{emnlp2021}

\usepackage{times}
\usepackage{latexsym}
\usepackage[T1]{fontenc}

\usepackage[utf8]{inputenc}
\usepackage{graphicx}
\usepackage[frozencache=true,cachedir=.]{minted}
\usepackage[]{xcolor}
\usepackage[inline]{enumitem}
\usepackage{scrextend}

\newcommand{\framework}{DomiKnowS}
\usepackage{microtype}

\usepackage{amsmath}
\DeclareMathOperator*{\argmax}{argmax} %
\title{\vspace*{-0.5in}
{{\small \hfill EMNLP Demo Track'2021}\\
\vspace*{.25in}}\framework: A Library for Integration of Symbolic Domain Knowledge in Deep Learning}

\author{Hossein Rajaby Faghihi$^1$, Quan Guo$^2$, Andrzej Uszok$^3$, Aliakbar Nafar$^1$, Elaheh Raisi$^1$, and \textbf{Parisa Kordjamshidi$^1$} \\
    $^1$ Michigan State University,
    $^2$ Sichuan University\\
  $^3$ Florida Institute for Human and Machine Cognition\\

  \small
  \texttt{
    rajabyfa@msu.edu,
    guoquan@scu.edu.cn,
    auszok@ihmc.org,
    } \\
    \small
    \texttt{
    \{nafarali, raisiela, kordjams\}@msu.edu} \\}

\begin{document}
\maketitle
\begin{abstract}
We demonstrate a library for the integration of domain knowledge in deep learning architectures. Using this library, the structure of the data is expressed symbolically via graph declarations and the logical constraints over outputs or latent variables can be seamlessly added to the deep models. The domain knowledge can be defined explicitly, which improves the models' explainability in addition to the performance and generalizability in the low-data regime. Several approaches for such an integration of symbolic and sub-symbolic models have been introduced; however, there is no library to facilitate the programming for such an integration in a generic way while various underlying algorithms can be used. 
Our library aims to simplify programming for such an integration in both training and inference phases while separating the knowledge representation from learning algorithms. We showcase various NLP benchmark tasks and beyond. The framework is publicly available at Github\footnote{https://github.com/HLR/DomiKnowS}.
\end{abstract}

\section{Introduction}
Current deep learning architectures are known to be data-hungry with issues mainly in generalizability and explainability~\cite{nguyen2015deep}. While these issues are hot research topics, one approach to address them is to inject external knowledge directly into the models when possible. 
While learning from examples revolutionized the way that intelligent systems are designed to gain knowledge, many tasks lack adequate data resources. Generating examples to capture knowledge is an expensive and lengthy process and especially not efficient when such a knowledge is available explicitly. Therefore, one main motivation of our \framework{} is to facilitate the integration of domain knowledge in deep learning architectures, in particular when this knowledge is represented symbolically. 

In this demonstration paper, we highlight the components of this framework that help to combine learning from data and exploiting knowledge in learning, including: 1) Learning problem specification 2) Knowledge representation 3) Algorithms for integration of knowledge and learning. Currently, \framework{} implementation relies on pyTorch and off-the-shelf optimization solvers such as Gurobi. However, it can be extended by developing hooks to other solvers and deep learning libraries since the interface is generic and independent from the underlying computational modules. 

In general, the integration of domain knowledge can be done 1) using pretrained models and transferring knowledge \cite{devlin2018bert,mirzaee2021spartqa}, 2) designing architectures that integrate knowledge expressed in knowledge bases (KB) and knowledge graphs (KG) in a way that the KB/KG context influences the learned representations \cite{yang2019leveraging,sun2018open}, or 3) using the knowledge explicitly and logically as a set of constrains or preferences over the inputs or outputs \cite{li2019augmenting,nandwani2019primal,muralidhar2018incorporating,stewart2017label}. Our current library aims at facilitating the third approach. While applying the constraints on input is technically trivial and could be done in a data pre-processing step, applying constraints over outputs and considering those structural constraints during training is a research challenge \cite{nandwani2019primal, li2019augmenting,inference-ijcai2020-382}. This requires encoding the knowledge at the algorithmic level. However, given that the constraints can be expressed logically and symbolically, having a language to express such a knowledge in a principled way is lacking in the current machine learning libraries. 
Using our developed \framework{} library, the domain knowledge will be provided symbolically and by the user utilizing a logical language that we have defined. This knowledge is used in various ways: a) As soft constraints by considering the violations as a part of loss function, this is done using a prim-dual formulation~\cite{nandwani2019primal} and can be expanded to probabilistic and sampling-based approaches~\cite{pmlr-v80-xu18h} or by mapping the constraint to differentiable operations~\cite{DBLP:conf/acl/LiS19} b)
mapping the constrains to an integer linear program and perform inference-based training by masking the loss~\cite{inference-ijcai2020-382}. Independent form the training paradigm the constraint can be always used as hard constraints during inference or not used at all.  

An interactive online demo of the framework is available at Google Colab\footnote{https://hlr.github.io/domiknows-nlp/} and the framework is accessible on GitHub\footnote{https://github.com/HLR/DomiKnowS}.

\section{Related Research}
Integration of domain knowledge in learning relates to tools that try to express the prior or posterior information about variables beyond what is in the data. This relates to probabilistic programming languages such as ~\cite{pfeffer16}, Venture~\cite{venture}, Stan~\cite{JSSv076i01}, and InferNet~\cite{InferNET}. The logical expression of domain knowledge is used in probabilistic logical programming languages such as ProbLog~\cite{Raedt07problog:a}, PRISM~\cite{SatoKa97}, the recent version of Problog, that is, Deep Problog~\cite{NEURIPS2018_dc5d637e}, Statistical Relational Learning tools, such as Markov logic networks~\cite{MLN}, Probabilistic soft logic~\cite{broecheler:uai10}, Bayesian Logic (BLOG)~\cite{MMRSOK05}, and slightly related to learning over graph structures~\cite{10.1145/3366424.3383111}.  Considering the structure of the output without its explicit declaration is considered in structured output prediction tools~\cite{rush-2020-torch}. This library is mostly related to the previous efforts for learning based programming and the integration of logical constraints in learning with classical machine learning approaches~\cite{citeulike:9883479,KordjamshidiRoWu15,KKCMSR16}. Our framework makes this connection to deep neural network libraries and arbitrarily designed architectures.
The unique feature of our library is that, the graph structure is defined symbolically based on the concepts in the domain. Despite Torch-struct~\cite{rush-2020-torch}, our library is independent from the underlying algorithms, and arbitrary structures can be expressed and used based on various underlying algorithms. In contrast to DeepProbLog, we are not limited to probabilistic inference and any solver can be used for inference depending on the training paradigm that is used for exploiting the logical constraints. Probabilistic soft logic is another framework that considers logical constraints in learning by mapping the constraint declarations to a Hing loss Markov random field~\cite{bac:jmlr17}. DRaiL is another declarative framework that is using logical constraints on top of deep learning and converts them to an integer linear program at at inference time~\cite{zhang-etal-2016-introducing}.   None of the above mentioned frameworks accommodate working with raw sensory data nor help in putting that in an operational structure that can form the domain predicates and be used by learning modules while our framework tries to address that challenge.
We support training paradigms that make use of the inference as a black box and in those cases any constraint optimization, logical inference engine or probabilistic inference tool can be integrated and used based on our abstraction and the provided modularity.  

\section{Declarative Learning-based Programming}

We use the Entity-Mention-Relation (EMR) extraction task to describe the framework. We discuss more showcases in Section \ref{sec:showcase}. 

{\bf Given} an input text 
 such as "\emph{Washington is employed by Associated Press.}", the task is to extract the entities and classify their types (e.g., people, organizations, and locations) as well as relations between them (e.g., works for, lives in). For example, for the above sentence \textit{[Washington]} is a ${person}$ \textit{ [Associated Press]} is an ${organization}$ and the relationship between these two entities is ${work}$-${for}$. 
We choose this task as it includes the prediction of multiple outputs at the sentence level, while there are global constraints over the outputs.\footnote{Please note this is just an example of a learning problem and does not have anything to do with the main functionality of the framework.}

In \framework{}, first, using our python-based specification language the user describes the problem and its logical constraints declarativly independent from the solutions. Second, it defines the necessary computational units (here PyTorch-based architectures) and connect the solution to the problem specification. Third, a program instance is created to execute the model using a background knowledge integration method with respect to the problem description.

\subsection{Problem Specification}
To model a problem in \framework{}, the user should specify the problem domain as a conceptual graph $\mathcal{G}(V,E)$. The nodes in $V$ represent concepts and the edges in $E$ are relationships. Each node can take a set of properties $P={P_1, P_2, ..., P_n}$.
Later, the logical constraints are expressed using the concepts in the graph. In EMR task, the graph contains some initial NLP concepts such as \textit{sentence}, \textit{phrase}, \textit{pair} and additional domain concepts such as \textit{people}, \textit{organization}, and \textit{work-for}.

\subsubsection{Concepts}
Each problem definition can contain three main types of concepts (nodes).

\noindent\textbf{Basic Concepts} define the structure of the input of the learning problem. For instance \textit{sentence}, \textit{phrase}, and \textit{word} are all base concepts that can be defined in the EMR task.

\noindent\textbf{Compositional Concepts} are used to define the many-to-many relationships between the basic concepts. Here, the \textit{pair} concept in the EMR task is a compositional concept. This is used as a basic concept for the relation extraction. We will further discuss this  when describing edges in Section \ref{sec:edges}.

\noindent\textbf{Decision Concepts} are derived concepts which are usually the outputs of the problem and subject to prediction. They are derived from the basic or compositional concepts. The \textit{people}, \textit{organization}, and \textit{work-for} are examples of derived concepts in the EMR conceptual graph.
Following is a partial snippet showing the definition of basic and compositional concepts for EMR task.
\begin{minted}[xleftmargin=1.5em,fontsize=\small,frame=lines,linenos,breaklines,escapeinside=!!]{python}
word = Concept(name='word')
phrase = Concept(name='phrase')
sentence = Concept(name='sentence')
pair = Concept(name='pair')
\end{minted}
The following snippet also shows the definition of some derived concepts in EMR example.
\begin{minted}[xleftmargin=1.5em,fontsize=\small,frame=lines,linenos,breaklines,escapeinside=!!]{python}
entity = phrase(name='entity')
people = entity(name='people')
org = entity(name='organization')
location = entity(name='location')
work_for = pair(name='work_for')
located_in = pair(name='located_in')
\end{minted}
The \textit{entity}, \textit{people}, \textit{organization} and \textit{location} are the derived concepts from the \textit{phrase} concept and the rest are derived from the \textit{pair}.

\subsubsection{Edges}
\label{sec:edges}
After defining the concepts, the user should specify existing relationships between them as edges in the conceptual graph.
Edges are used to either map instances from one concept to another, or generate instances of a concept from another concept. \framework{} only supports a set of predefined edge types, namely \textit{is\_a}, \textit{has\_a}, \textit{contains},. 
\noindent\textbf{is\_a} is automatically defined between a derived concept and its parent. In the EMR example, there is an \textit{is\_a} edge between \textit{people} and \textit{entity}. 
\textit{is\_a} is mostly used to introduce hierarchical constraints and relate the basic and derived concepts.

\noindent\textbf{Has\_a} connects a compositional concept to its components (also referred to as \textit{arguments}). In the EMR example, \textit{pair} concept has two \textit{has\_a} edges to the \textit{phrase} concept to specify the \textit{arg1} and \textit{arg2} of the composition. We allow an arbitrary number of arguments in a \textit{has\_a} relationship, see below.
\begin{minted}[xleftmargin=1em,fontsize=\small,frame=lines,linenos,breaklines,escapeinside=!!]{python}
pair.has_a(arg1=phrase, arg2=phrase)
\end{minted}
\noindent\textbf{Contains} edge defines a one-to-many relationship between two concepts to represent (parent, child) relationship. Here, the parents of a concept is not necessarily limited to be only one. Following is a sample snippet to define a \textit{contains} edge between \textit{sentence} and \textit{phrase}:
\begin{minted}[xleftmargin=1em,fontsize=\small,frame=lines,linenos,breaklines,escapeinside=!!]{python}
sentence.contains(phrase)
\end{minted}


\subsubsection{Global Constraints}
\label{section:gc}
The constraint definition is the part where the prior knowledge of the problem is defined to enable domain integration. The constraints of each task should be defined on top of the problem using the specified concepts and relationships there.

The constraints can be 1) automatically inferred from the conceptual graph structure, 2) extracted from the standard ontology formalism (here OWL\footnote{Ontology Web Language}), 3) explicitly defined in the \framework{}'s logical constraint language. The framework internally uses defined constraints in the training-time or inference-time optimization depending on the integration method selected for the task. We discuss the inference phase in more details in section \ref{sec:inference}.
Here is an example of constraint in our constraint language for the EMR task:
\begin{minted}[xleftmargin=1.5em,fontsize=\small,frame=lines,linenos,breaklines,escapeinside=!!]{python}
ifL(work_for('x'), andL(people(path= ('x',arg1)), organization(path='x',arg2)))
\end{minted}
The above constraint indicates that a \textit{work\_for} relationship only holds between \textit{people} and \textit{organization}. Other syntactic variations of this constraint are shown in the Appendix.

To process constraints, \framework{} maps those to a set of equivalent algebraic inequalities or their soft logic interpretation depending on the integration method. We discuss this more in Section \ref{sec:inference}.



\subsection{Model Declaration}
\label{sec:model}
Model declaration phase is about defining the computational units of the task. The basic building blocks of the model in \framework{} are \textit{sensors} and \textit{learners}, which are used to define either deterministic or probabilistic functionalities of the model. Sensor/Learners interact with the conceptual graph by defining properties on the concepts (nodes). Each sensor/learner receives a set of inputs either from the raw data or property values on the graph and introduces new property values. Sensors are computational units with no trainable parameters; and learners are the ones including the neural models. As stated before, the model declaration phase only defines the connection of the graph properties to the computational units and the execution is done later by the program instances.




The user can use any deep learning architecture compatible with pyTorch modules alongside the set of pre-designed and commonly-used neural architectures currently existing in the framework.
To facilitate modeling different architectures and computational algorithms in 
\framework{}, we provide a set of predefined sensors to do basic mathematical operations and linguistic feature extraction. 
Following is a short snippet of defining some sensors/learners for the EMR task.
\begin{minted}[xleftmargin=0em,fontsize=\small,frame=lines,linenos,breaklines,escapeinside=!!]{python}
phrase['w2v'] = FunctionalSensor('text', forward=word2vec)
phrase[people] = ModuleLearner('w2v', module=Classifier(FEATURE_DIM))
pair[work_for] = ModuleLearner('emb', module=Classifier(FEATURE_DIM*2))
\end{minted}
In this example, the sensor \textit{Word2Vec} is used to obtain token representations from the ``text'' property of each \textit{phrase}. There is also a very simple and straightforward linear neural model to classify \textit{phrases} and \textit{pairs} into different classes such as \textit{people}, \textit{organization}, etc.



\section{Learning and Evaluation in \framework{}}
\label{sec:program}
To execute the defined model with respect to the problem graph, \framework{} uses program instances. A program instance is responsible to run the model, apply loss functions, optimize the parameters, connect the output decisions to the inference algorithms, and generate the final results and metrics. 
Executing the program instance relies on the problem graph, model declaration, dataloaders and a backbone data structure called DataNode. \textbf{DataLoader} provides an iterable object to loop over the data. \textbf{DataNode} is an instance of the conceptual graph to keep track of the data instances and store the computational results of the sensors and learners. For the EMR task the program definition is as follows:
\begin{minted}[xleftmargin=1.5em,fontsize=\small,frame=lines,linenos,breaklines,escapeinside=!!]{python}
program = Program(graph, poi=(sentence, phrase, pair), loss=NBCrossEntropyLoss(), metric=PRF1())
\end{minted}
Here, the concepts passed to the \textit{poi} field specifies the training points of the program. This enables the user to only train parts of the defined model. 

For each program instance, the user should specify the domain knowledge integration method. The available methods for integration is discussed in Section \ref{sec:inference}. After initializing the program, the user can call \textit{train}, \textit{test}, and \textit{prediction} functionalities to train and evaluate the designed model. 
Below snippet is to run training and evaluation on the EMR task:
\begin{minted}[xleftmargin=1.5em,fontsize=\small,frame=lines,linenos,breaklines,escapeinside=!!]{python}
program.train(train_reader, test_reader, epochs=10, Optim=torch.optim.SGD(param, lr=.001))
program.test(new_test_reader)
\end{minted}
Here, the user will specify the dataloaders for different sets of the data and the hyper-parameters required to train the model.

Programs can be composed to address different training paradigms such as end-to-end or pipeline training by defining different training points for each program. More details available in the Appendix.
Following is an alternative program definition for the pre-training phrases and training pairs:
\begin{minted}[xleftmargin=1.5em,fontsize=\small,frame=lines,linenos,breaklines,escapeinside=!!]{python}
program_1 = Program(graph, poi=(phrase, sentence))
program_2 = Program(graph, poi=(pair))
program_1.train(); program_2.train()
\end{minted}

\subsection{Inference and Optimization}
\label{sec:inference}
\framework{} provides access to a set of approaches to integrate background knowledge in the form of constraints on the output decisions or latent variables/concepts.
Currently, \framework{} addresses three different paradigms for integration: 1) Learning + prediction time inference (L+I) 2) Training-time integration with hard constraints 3) Training-time integration with soft constraints.
The first method, which we refer to as enforcing global constraints can also be combined and applied on top of the second and third approaches at inference-time.


\noindent\textbf{Prediction-time Inference}:
In the back-end of \framework{}, ILP~\footnote{Integer Linear Programming}~\cite{RothYi05} solvers are used to make inference under global linear constraints. The constraints are denoted by $\mathcal{C}\left(\cdot\right) \le 0$. Without loss of generality, we can denote the structured output as a binary vector $y\in\mathcal{R}^n$. Given local predictions $F(\theta)$ from the neural network, the global inference can be modeled to maximize the combination of log probability scores subject to the constraints \cite{RothYi05,inference-ijcai2020-382} as follows,
\begin{equation}
\label{eq:ilpv}
\begin{aligned}
F^*(\theta) &= \argmax_{y} \log{F(\theta)}^\top y \\
            & \mbox{subject to} \quad \mathcal{C}\left(y\right)\le0.
\end{aligned}
\end{equation}
To handle constraints in ILP, we create variables for each local decision of instances and transform the logical constraints to algebraic inequalities~\cite{citeulike:9883479} in terms of those variables. Auxiliary variables are added to represent the nested constraints.
The inference method can be extended to support other approaches such as probabilistic inference and dynamic programming in future.

\noindent\textbf{Integration of hard constraint in training:}
Here, we use our proposed inference-masked loss approach (IML)~\cite{inference-ijcai2020-382} which constructs a mask over local predictions based on the global inference results. The main intuition is to avoid updating the model based on local violations when the global inference can recover true labels from the current predictions.
Given structured prediction $F(\theta)$ from a neural network and its global inference $F^*(\theta)$ subject to the constraints, IML is extended from negative log likelihood as follows
\begin{equation}
    \label{eq:imlv}
    \begin{split}
        &\mathcal{L}_{\mbox{IML}}\left(F(\theta),Y\right) =\\& - \left( \left(1 - F^*(\theta)\right) \odot Y \right)^\top \log{F(\theta)},
    \end{split}
\end{equation}
where $Y$ is the structured ground-truth labels and $\odot$ indicates element-wise product.
We implemented $\mathcal{L}_{\mbox{IML}\left(\lambda\right)}$ which balances between negative log likelihood and IML with a factor $\lambda$ as in \cite{inference-ijcai2020-382}.
IML works best for very low-resource tasks where label disambiguation cannot be learned from the data but easier to be done given available relational constraints between instances of training samples.
The required constraint mapping for the IML is the same as the global constraint optimization tool (here ILP).

\noindent\textbf{Integration of soft constraints in training:}
We use the primal-dual formulation of constraints proposed in \cite{Nandwani2019APD} to integrate soft constraints in training the models.
Primal-Dual considers the constraints in the neural network training by augmenting the loss function using Lagrangian multipliers $\Lambda$ for the violations of the constraints by the set of predictions. The constraints are regularized by a hinge function $\left[\mathcal{C}\left(F\left(\theta\right)\right)\right]^{+}$. The problem is formulated as a min-max optimization where it maximizes the Lagrangian function with the multipliers to enforce the constraints and minimize it with the parameters in the neural network. Here, instead of solving the min-max primal, we solve the max-min dual of the original problem.
\begin{equation}
    \max_{\Lambda}\min_{\theta} \mathcal{L}\left(F\left(\theta\right),Y\right) + \Lambda^{\top} \left[\mathcal{C}\left(F\left(\theta\right)\right)\right]^{+}.
\end{equation}
During training, we optimize by minimization and maximization alternatively.
With Primal-Dual strategy, the model learns to obey the constraints without the help from additional inference.
Primal-Dual is less time-consuming at prediction-time than the previous methods as it does not need an additional ILP optimization phase. This can also be used for semi-supervised setting using domain knowledge.
Handling constraints in Primal-Dual is done by mapping them to their respective soft logic interpretation \cite{nandwani2019primal}.

It is an open research topic to identify which of the integration methods performs best for different tasks. However, \framework{} makes it effortless to use one problem specification and run all the aforementioned methods.

\section{Showcases}
\label{sec:showcase}
The effectiveness of ILP~\cite{RothYi05}, IML~\cite{inference-ijcai2020-382}, and Primal-Dual~\cite{Nandwani2019APD} methods have been already shown in their respective papers. Here, we provide different tasks and settings to showcase our framework's abilities and flexibility to model various problems.
The results, models' implementation and details of experiments are (partially) available in the Supplementary part of this paper and (completely) in \framework{}' GitHub Repository\footnote{\label{note:github}https://github.com/HLR/DomiKnowS}. 
\subsection{EMR}
Our implementation of the EMR task is based on the CoNLL~\cite{sang2003introduction} benchmark and follows the same setting as in \cite{inference-ijcai2020-382}. The model uses BERT~\cite{devlin-etal-2019-bert} for token representation and a linear boolean classifier for each derived concept. 
The constraints used in this experiment are the domain and range constraints of pairs and the mutual exclusiveness of different derived concepts, which were seen during the previous sections. 
IML and Primal-Dual methods perform the same as the baseline using both 100\% and 25\% of the data, while ILP inference achieves 1.3\% improvement on the 100\% of data and 0.6\% improvement on 25\% of the data. More details are available in the Appendix.

\subsection{Question Answering}
We use WIQA~\cite{tandon-etal-2019-wiqa} benchmark as a sample question answering task in \framework{}. The problem graph contains \textit{paragraph}, \textit{question}, \textit{symmetric}, and \textit{transitive} concepts. Each \textit{paragraph} comes with a set of \textit{question}s. As explained by \citeauthor{asai2020logicguided}, enforcing constraints between different question answers is beneficial to the models' performance. By modeling those constraints in \framework{}, ILP improves the model's accuracy from 74.22\% to 79.05\%, IML reaches 75.49\%, Primal-Dual achieves 76.59\%, and the combination of Primal-Dual and ILP performs best with 80.35\% accuracy. More details are available in the Appendix.
Following is a sample constraint defined for the task.
\begin{minted}[xleftmargin=1.5em,fontsize=\small,frame=lines,linenos,breaklines,escapeinside=!!]{python}
symmetric.has_a(arg1=question, arg2=question)
ifL(is_more('x'), is_less(path=('x', arg2))
\end{minted}
\subsection{Image Classification}
We use CIFAR-10 benchmark ~\cite{cifar10} to show image classification task in \framework{}. CIFAR-10 consists of 60,000 colourful images of 10 classes with 6,000 image for each class. To construct the graph, we defined the derived concepts, \textit{airplane}, \textit{dog}, \textit{truck}, \textit{automobile}, \textit{bird}, \textit{cat}, \textit{deer}, \textit{frog}, \textit{horse} and \textit{ship}, and the base concept \textit{image}.
% ...
We introduce the disjoint constraint between the labels of an image. For this problem, hierarchical constraints between class labels can also be used easily. 
\begin{minted}[xleftmargin=1em,fontsize=\small,frame=lines,linenos,breaklines,escapeinside=!!]{python}
disjoint(truck, dog, airplane, automobile, bird, cat, deer, frog, horse, ship)
\end{minted}

Both the disjoint and hierarchical constraints do not affect the accuracy of the task by a large margin and ILP can only achieve near 0.5\% improvement over the classification task. 


\subsection{Inference-Only Example}
This example is to show that \framework{} can solve pure optimization problems as well. The task is similar to the classic graph-coloring problem. A set of cities are given each of which can have a fire station or not. We want to allocate the fire stations to cities in a way that the following constraint is met.

\noindent\textbf{Constraint: }
For each city $x$, it either has a fire station or there exists a city $y$ which is a neighbor of city $x$ and has a fire station.

To implement this we define the basic concept \textit{city}, the \textit{neighbor} relationship between two cities and the derived concept \textit{FirestationCity}.
\begin{minted}[xleftmargin=1.5em,fontsize=\small,frame=lines,linenos,breaklines,escapeinside=!!]{python}
neighbor.has_a(arg1=city, arg2=city)
orL(firestationCity('x'), existsL(firestationCity(path=('x', neighbor.arg2))
\end{minted}

We have also included more showcases to solve sentiment analysis~\cite{go2009twitter} and email spam detection in our GitHub repository~\footref{note:github}. We will add models for procedural reasoning~\cite{faghihi2021timestamped} and spatial role labeling~\cite{mirzaee2021spartqa} in future. 


\section{Conclusions and Future Work}
\framework{} makes it effortless to integrate domain knowledge into deep neural network models using a unified framework. It allows users to switch between different algorithms and benefit from a rich source of abstracted functionalities and computational modules developed for multiple tasks. It allows naming concepts, defining their relationships symbolically and combining symbolic and sub-symbolic reasoning over the named concepts. \framework{} helps in interpretability of neural architectures by providing named layers and access to the neural models' computations at each stage of the training and evaluation process. As a future direction, we are looking to enrich our library with predefined functionalities and neural models and further extend the framework's ability to support more methods on integration of domain knowledge with deep neural models as well as seamless model composition. \framework{} is publicly available at our website\footnote{https://hlr.github.io/domiknows-nlp/} and on GitHub.\footnote{https://github.com/HLR/DomiKnowS}

\section*{Acknowledgements}

This project is partially funded by the Office of Naval Research~(ONR) grant \texttt{\#}N00014-20-1-2005. 

\bibliography{anthology.bib,acl2021.bib,allpapers.bib,ccg.bib,new.bib,ijcai18.bib,cited.bib,LBP_references.bib,QAbib.bib}

\begin{thebibliography}{38}
\expandafter\ifx\csname natexlab\endcsname\relax\def\natexlab#1{#1}\fi

\bibitem[{Asai and Hajishirzi(2020)}]{asai2020logicguided}
Akari Asai and Hannaneh Hajishirzi. 2020.
\newblock \href {http://arxiv.org/abs/2004.10157} {Logic-guided data
  augmentation and regularization for consistent question answering}.

\bibitem[{Bach et~al.(2017)Bach, Broecheler, Huang, and Getoor}]{bac:jmlr17}
Stephen~H. Bach, Matthias Broecheler, Bert Huang, and Lise Getoor. 2017.
\newblock \href {https://github.com/stephenbach/bach-jmlr17-code} {Hinge-loss
  markov random fields and probabilistic soft logic}.
\newblock \emph{Journal of Machine Learning Research (JMLR)}, 18:1--67.

\bibitem[{Broecheler et~al.(2010)Broecheler, Mihalkova, and
  Getoor}]{broecheler:uai10}
Matthias Broecheler, Lilyana Mihalkova, and Lise Getoor. 2010.
\newblock Probabilistic similarity logic.
\newblock In \emph{Conference on Uncertainty in Artificial Intelligence}.

\bibitem[{Carpenter et~al.(2017)Carpenter, Gelman, Hoffman, Lee, Goodrich,
  Betancourt, Brubaker, Guo, Li, and Riddell}]{JSSv076i01}
Bob Carpenter, Andrew Gelman, Matthew Hoffman, Daniel Lee, Ben Goodrich,
  Michael Betancourt, Marcus Brubaker, Jiqiang Guo, Peter Li, and Allen
  Riddell. 2017.
\newblock Stan: A probabilistic programming language.
\newblock \emph{Journal of Statistical Software, Articles}, 76(1):1--32.

\bibitem[{{De Raedt} et~al.(2007){De Raedt}, Kimmig, and
  Toivonen}]{Raedt07problog:a}
Luc {De Raedt}, Angelika Kimmig, and Hannu Toivonen. 2007.
\newblock Problog: a probabilistic {P}rolog and its application in link
  discovery.
\newblock In \emph{Proceedings of the 20th International Joint Conference on
  Artificial Intelligence}, pages 2468--2473. AAAI Press.

\bibitem[{Devlin et~al.(2018)Devlin, Chang, Lee, and
  Toutanova}]{devlin2018bert}
Jacob Devlin, Ming-Wei Chang, Kenton Lee, and Kristina Toutanova. 2018.
\newblock Bert: Pre-training of deep bidirectional transformers for language
  understanding.
\newblock \emph{arXiv preprint arXiv:1810.04805}.

\bibitem[{Devlin et~al.(2019)Devlin, Chang, Lee, and
  Toutanova}]{devlin-etal-2019-bert}
Jacob Devlin, Ming-Wei Chang, Kenton Lee, and Kristina Toutanova. 2019.
\newblock \href {https://doi.org/10.18653/v1/N19-1423} {{BERT}: Pre-training of
  deep bidirectional transformers for language understanding}.
\newblock In \emph{Proceedings of the 2019 Conference of the North {A}merican
  Chapter of the Association for Computational Linguistics: Human Language
  Technologies, Volume 1 (Long and Short Papers)}, pages 4171--4186,
  Minneapolis, Minnesota. Association for Computational Linguistics.

\bibitem[{Domingos and Richardson(2004)}]{MLN}
Perdo Domingos and Matthew Richardson. 2004.
\newblock Markov logic: A unifying framework for statistical relational
  learning.
\newblock In \emph{{ICML}'04 Workshop on Statistical Relational Learning and
  its Connections to Other Fields}, pages 49--54.

\bibitem[{Faghihi and Kordjamshidi(2021)}]{faghihi2021timestamped}
Hossein~Rajaby Faghihi and Parisa Kordjamshidi. 2021.
\newblock \href {http://arxiv.org/abs/2104.07635} {Time-stamped language model:
  Teaching language models to understand the flow of events}.

\bibitem[{Go et~al.(2009)Go, Huang, and Bhayani}]{go2009twitter}
Alec Go, Lei Huang, and Richa Bhayani. 2009.
\newblock Twitter sentiment analysis.

\bibitem[{Guo et~al.(2020)Guo, Rajaby~Faghihi, Zhang, Uszok, and
  Kordjamshidi}]{inference-ijcai2020-382}
Quan Guo, Hossein Rajaby~Faghihi, Yue Zhang, Andrzej Uszok, and Parisa
  Kordjamshidi. 2020.
\newblock \href {https://doi.org/10.24963/ijcai.2020/382} {Inference-masked
  loss for deep structured output learning}.
\newblock In \emph{Proceedings of the Twenty-Ninth International Joint
  Conference on Artificial Intelligence, {IJCAI-20}}, pages 2754--2761.
  International Joint Conferences on Artificial Intelligence Organization.
\newblock Main track.

\bibitem[{Kordjamshidi et~al.(2015)Kordjamshidi, Roth, and
  Wu}]{KordjamshidiRoWu15}
P.~Kordjamshidi, D.~Roth, and H.~Wu. 2015.
\newblock \href {http://cogcomp.cs.illinois.edu/papers/KordjamshidiRoWu15.pdf}
  {Saul: Towards declarative learning based programming}.
\newblock In \emph{Proc. of the International Joint Conference on Artificial
  Intelligence (IJCAI)}.

\bibitem[{Kordjamshidi et~al.(2016)Kordjamshidi, Khashabi, Christodoulopoulos,
  Mangipudi, Singh, and Roth}]{KKCMSR16}
Parisa Kordjamshidi, Daniel Khashabi, Christos Christodoulopoulos, Bhargav
  Mangipudi, Sameer Singh, and Dan Roth. 2016.
\newblock \href {http://cogcomp.org/papers/KKCMSR16.pdf} {Better call saul:
  Flexible programming for learning and inference in nlp}.
\newblock In \emph{Proc. of the International Conference on Computational
  Linguistics (COLING)}.

\bibitem[{Krizhevsky et~al.()Krizhevsky, Nair, and Hinton}]{cifar10}
Alex Krizhevsky, Vinod Nair, and Geoffrey Hinton.
\newblock \href {http://www.cs.toronto.edu/~kriz/cifar.html} {Cifar-10
  (canadian institute for advanced research)}.

\bibitem[{Li and Srikumar(2019{\natexlab{a}})}]{li2019augmenting}
Tao Li and Vivek Srikumar. 2019{\natexlab{a}}.
\newblock Augmenting neural networks with first-order logic.
\newblock \emph{arXiv preprint arXiv:1906.06298}.

\bibitem[{Li and Srikumar(2019{\natexlab{b}})}]{DBLP:conf/acl/LiS19}
Tao Li and Vivek Srikumar. 2019{\natexlab{b}}.
\newblock \href {https://doi.org/10.18653/v1/p19-1028} {Augmenting neural
  networks with first-order logic}.
\newblock In \emph{Proceedings of the 57th Conference of the Association for
  Computational Linguistics, {ACL} 2019, Florence, Italy, July 28- August 2,
  2019, Volume 1: Long Papers}, pages 292--302. Association for Computational
  Linguistics.

\bibitem[{Manhaeve et~al.(2018)Manhaeve, Dumancic, Kimmig, Demeester, and
  De~Raedt}]{NEURIPS2018_dc5d637e}
Robin Manhaeve, Sebastijan Dumancic, Angelika Kimmig, Thomas Demeester, and Luc
  De~Raedt. 2018.
\newblock \href
  {https://proceedings.neurips.cc/paper/2018/file/dc5d637ed5e62c36ecb73b654b05ba2a-Paper.pdf}
  {Deepproblog: Neural probabilistic logic programming}.
\newblock In \emph{Advances in Neural Information Processing Systems},
  volume~31. Curran Associates, Inc.

\bibitem[{Mansinghka et~al.(2014)Mansinghka, Selsam, and Perov}]{venture}
Vikash~K. Mansinghka, Daniel Selsam, and Yura~N. Perov. 2014.
\newblock \href {http://arxiv.org/abs/1404.0099} {Venture: a higher-order
  probabilistic programming platform with programmable inference}.
\newblock \emph{CoRR}, abs/1404.0099.

\bibitem[{Milch et~al.(2005)Milch, Marthi, Russell, Sontag, Ong, and
  Kolobov}]{MMRSOK05}
Brian Milch, Bhaskara Marthi, Stuart Russell, David Sontag, Daniel~L. Ong, and
  Andrey Kolobov. 2005.
\newblock \href {http://www.cs.berkeley.edu/~russell/papers/ijcai05-blog.pdf}
  {{BLOG}: Probabilistic models with unknown objects}.
\newblock In \emph{Proceedings of the International Joint Conference on
  Artificial Intelligence (IJCAI)}.

\bibitem[{Minka et~al.(2012)Minka, Winn, Guiver, and Knowles}]{InferNET}
Tom Minka, John~M. Winn, John~P. Guiver, and David~A. Knowles. 2012.
\newblock {Infer.NET 2.5}.
\newblock Microsoft Research Cambridge. http://research.microsoft.com/infernet.

\bibitem[{Mirzaee et~al.(2021)Mirzaee, Faghihi, Ning, and
  Kordjamshidi}]{mirzaee2021spartqa}
Roshanak Mirzaee, Hossein~Rajaby Faghihi, Qiang Ning, and Parisa Kordjamshidi.
  2021.
\newblock Spartqa: A textual question answering benchmark for spatial
  reasoning.
\newblock In \emph{Proceedings of the 2021 Conference of the North American
  Chapter of the Association for Computational Linguistics: Human Language
  Technologies}, pages 4582--4598.

\bibitem[{Muralidhar et~al.(2018)Muralidhar, Islam, Marwah, Karpatne, and
  Ramakrishnan}]{muralidhar2018incorporating}
Nikhil Muralidhar, Mohammad~Raihanul Islam, Manish Marwah, Anuj Karpatne, and
  Naren Ramakrishnan. 2018.
\newblock Incorporating prior domain knowledge into deep neural networks.
\newblock In \emph{2018 IEEE International Conference on Big Data (Big Data)},
  pages 36--45. IEEE.

\bibitem[{Nandwani et~al.(2019{\natexlab{a}})Nandwani, Pathak, Mausam, and
  Singla}]{Nandwani2019APD}
Yatin Nandwani, Abhishek Pathak, Mausam, and Parag Singla. 2019{\natexlab{a}}.
\newblock A primal dual formulation for deep learning with constraints.
\newblock In \emph{NeurIPS}.

\bibitem[{Nandwani et~al.(2019{\natexlab{b}})Nandwani, Pathak, Singla
  et~al.}]{nandwani2019primal}
Yatin Nandwani, Abhishek Pathak, Parag Singla, et~al. 2019{\natexlab{b}}.
\newblock A primal dual formulation for deep learning with constraints.

\bibitem[{Nguyen et~al.(2015)Nguyen, Yosinski, and Clune}]{nguyen2015deep}
Anh Nguyen, Jason Yosinski, and Jeff Clune. 2015.
\newblock Deep neural networks are easily fooled: High confidence predictions
  for unrecognizable images.
\newblock In \emph{Proceedings of the IEEE conference on computer vision and
  pattern recognition}, pages 427--436.

\bibitem[{Pfeffer(2016)}]{pfeffer16}
Avi Pfeffer. 2016.
\newblock \emph{Practical Probabilistic Programming}.
\newblock Manning Publications.

\bibitem[{Rizzolo and Roth(2010)}]{citeulike:9883479}
N.~Rizzolo and D.~Roth. 2010.
\newblock Learning based {J}ava for rapid development of {NLP} systems.
\newblock In \emph{Proceedings of the Seventh Conference on International
  Language Resources and Evaluation}.

\bibitem[{Roth and Yih(2005)}]{RothYi05}
D.~Roth and W.~Yih. 2005.
\newblock \href {http://cogcomp.org/papers/RothYi05.pdf} {Integer linear
  programming inference for conditional random fields}.
\newblock In \emph{Proc. of the International Conference on Machine Learning
  (ICML)}, pages 737--744.

\bibitem[{Rush(2020)}]{rush-2020-torch}
Alexander Rush. 2020.
\newblock \href {https://doi.org/10.18653/v1/2020.acl-demos.38} {Torch-struct:
  Deep structured prediction library}.
\newblock In \emph{Proceedings of the 58th Annual Meeting of the Association
  for Computational Linguistics: System Demonstrations}, pages 335--342,
  Online. Association for Computational Linguistics.

\bibitem[{Sang and De~Meulder(2003)}]{sang2003introduction}
Erik Tjong~Kim Sang and Fien De~Meulder. 2003.
\newblock Introduction to the conll-2003 shared task: Language-independent
  named entity recognition.
\newblock In \emph{Proceedings of the Seventh Conference on Natural Language
  Learning at HLT-NAACL 2003}, pages 142--147.

\bibitem[{Sato and Kameya(1997)}]{SatoKa97}
Taisuke Sato and Yoshitaka Kameya. 1997.
\newblock Prism: A language for symbolic-statistical modeling.
\newblock In \emph{Proceedings of the International Joint Conference on
  Artificial Intelligence (IJCAI)}, pages 1330--1339.

\bibitem[{Stewart and Ermon(2017)}]{stewart2017label}
Russell Stewart and Stefano Ermon. 2017.
\newblock Label-free supervision of neural networks with physics and domain
  knowledge.
\newblock In \emph{Proceedings of the AAAI Conference on Artificial
  Intelligence}, volume~31.

\bibitem[{Sun et~al.(2018)Sun, Dhingra, Zaheer, Mazaitis, Salakhutdinov, and
  Cohen}]{sun2018open}
Haitian Sun, Bhuwan Dhingra, Manzil Zaheer, Kathryn Mazaitis, Ruslan
  Salakhutdinov, and William~W Cohen. 2018.
\newblock Open domain question answering using early fusion of knowledge bases
  and text.
\newblock \emph{arXiv preprint arXiv:1809.00782}.

\bibitem[{Tandon et~al.(2019)Tandon, Dalvi, Sakaguchi, Clark, and
  Bosselut}]{tandon-etal-2019-wiqa}
Niket Tandon, Bhavana Dalvi, Keisuke Sakaguchi, Peter Clark, and Antoine
  Bosselut. 2019.
\newblock \href {https://doi.org/10.18653/v1/D19-1629} {{WIQA}: A dataset for
  {``}what if...{''} reasoning over procedural text}.
\newblock In \emph{Proceedings of the 2019 Conference on Empirical Methods in
  Natural Language Processing and the 9th International Joint Conference on
  Natural Language Processing (EMNLP-IJCNLP)}, pages 6076--6085, Hong Kong,
  China. Association for Computational Linguistics.

\bibitem[{Xu et~al.(2018)Xu, Zhang, Friedman, Liang, and Van~den
  Broeck}]{pmlr-v80-xu18h}
Jingyi Xu, Zilu Zhang, Tal Friedman, Yitao Liang, and Guy Van~den Broeck. 2018.
\newblock \href {https://proceedings.mlr.press/v80/xu18h.html} {A semantic loss
  function for deep learning with symbolic knowledge}.
\newblock In \emph{Proceedings of the 35th International Conference on Machine
  Learning}, volume~80 of \emph{Proceedings of Machine Learning Research},
  pages 5502--5511. PMLR.

\bibitem[{Yang and Mitchell(2019)}]{yang2019leveraging}
Bishan Yang and Tom Mitchell. 2019.
\newblock Leveraging knowledge bases in lstms for improving machine reading.
\newblock \emph{arXiv preprint arXiv:1902.09091}.

\bibitem[{Zhang et~al.(2016)Zhang, Pacheco, Li, and
  Goldwasser}]{zhang-etal-2016-introducing}
Xiao Zhang, Maria~Leonor Pacheco, Chang Li, and Dan Goldwasser. 2016.
\newblock \href {https://doi.org/10.18653/v1/W16-5906} {Introducing {DRAIL}
  {--} a step towards declarative deep relational learning}.
\newblock In \emph{Proceedings of the Workshop on Structured Prediction for
  {NLP}}, pages 54--62, Austin, TX. Association for Computational Linguistics.

\bibitem[{Zheng et~al.(2020)Zheng, Wang, Gan, Zhang, and
  Karypis}]{10.1145/3366424.3383111}
Da~Zheng, Minjie Wang, Quan Gan, Zheng Zhang, and George Karypis. 2020.
\newblock \href {https://doi.org/10.1145/3366424.3383111} {Learning graph
  neural networks with deep graph library}.
\newblock WWW '20, New York, NY, USA. Association for Computing Machinery.

\end{thebibliography}
\bibliographystyle{acl_natbib}

\clearpage
\appendix
\section{Global Constraints and Mapping}
Following is an example of the mapping between OWL constraint, graph structure and the logic python constraint. 
The ontology definition in OWL:
\begin{minted}[xleftmargin=1em,fontsize=\small,frame=lines,linenos,breaklines,escapeinside=!!]{python}
<owl:ObjectProperty rdf:ID="work_for">
    <rdfs:domain rdf:resource="#people"/>
    <rdfs:range rdf:resource="#organization"/>
</owl:ObjectProperty>
\end{minted}
or equivalent graph structure definition:
\begin{minted}[xleftmargin=1em,fontsize=\small,frame=lines,linenos,breaklines,escapeinside=!!]{python}
work_for.has_a(arg1=people, arg2=organization)
\end{minted}
\framework{}'s constrain language representation:
\begin{minted}[xleftmargin=1.5em,fontsize=\small,frame=lines,linenos,breaklines,escapeinside=!!]{python}
ifL(work_for('x'), andL(people(path= ('x',arg1)), organization(path='x',arg2)))
\end{minted}

All three above constraints represent the same knowledge that a \textit{work\_for} relationship only holds between \textit{people} and \textit{organization}.

In order to map this logical constrain to ILP, the solver collects sets of candidates for each used in the constrain concepts. 

ILP inequalities are created for each of the combinations of candidates sets.  The internal nested \textit{andL} logical expression is translated to a set of three algebraic inequalities. The new variable \textit{varAND}) is created to transfer the result of the internal expression into the external one.
\begin{minted}[xleftmargin=1em,fontsize=\small,frame=lines,linenos,breaklines,escapeinside=!!]{python}
varAND <= varPhraseIsPeople
varAND <= varPhraseIsOrganization
varPhraseIsPeople + varPhraseIsOrganization <= varAND + 1
\end{minted}

External \textit{ifL} expression is translated to a single algebraic inequality (refers to the variable \textit{varAND}):
\begin{minted}[xleftmargin=1em,fontsize=\small,frame=lines,linenos,breaklines,escapeinside=!!]{python}
varPhraseIsWorkFor <= varAND
\end{minted}

\section{Program Composition}
The Program instances allow the user to define different training tasks without extra effort to change the underlying models. One can define end-to-end, pipelines, and pre-training and fine-tuning steps just by defining different program instances and calling them one after another. 
For instance, we can seamlessly switch between the following variations of learning paradigms on the EMR task.

\noindent End-To-End training:
\begin{minted}[xleftmargin=1.5em,fontsize=\small,frame=lines,linenos,breaklines,escapeinside=!!]{python}
program = Program(graph, poi=(phrase, sentence, pair))
program.train()
\end{minted}
Pre-train phrase then just train pair:
\begin{minted}[xleftmargin=1.5em,fontsize=\small,frame=lines,linenos,breaklines,escapeinside=!!]{python}
program_1 = Program(graph, poi=(phrase, sentence))
program_2 = Program(graph, poi=(pair))
program_1.train(); program_2.train()
\end{minted}

Pre-train phrase and use the result in the end-to-end training:
\begin{minted}[xleftmargin=1.5em,fontsize=\small,frame=lines,linenos,breaklines,escapeinside=!!]{python}
program_1 = POIProgram(graph, poi=(phrase, sentence), ...)
program_2 = POIProgram(graph, poi=(phrase, sentence, pair), ...)
program_1.train(...)
program_2.train(...)
\end{minted}

\section{Experiments}

\subsection{EMR}
\begin{figure*}[]
    \centering
    \includegraphics[width=1\textwidth]{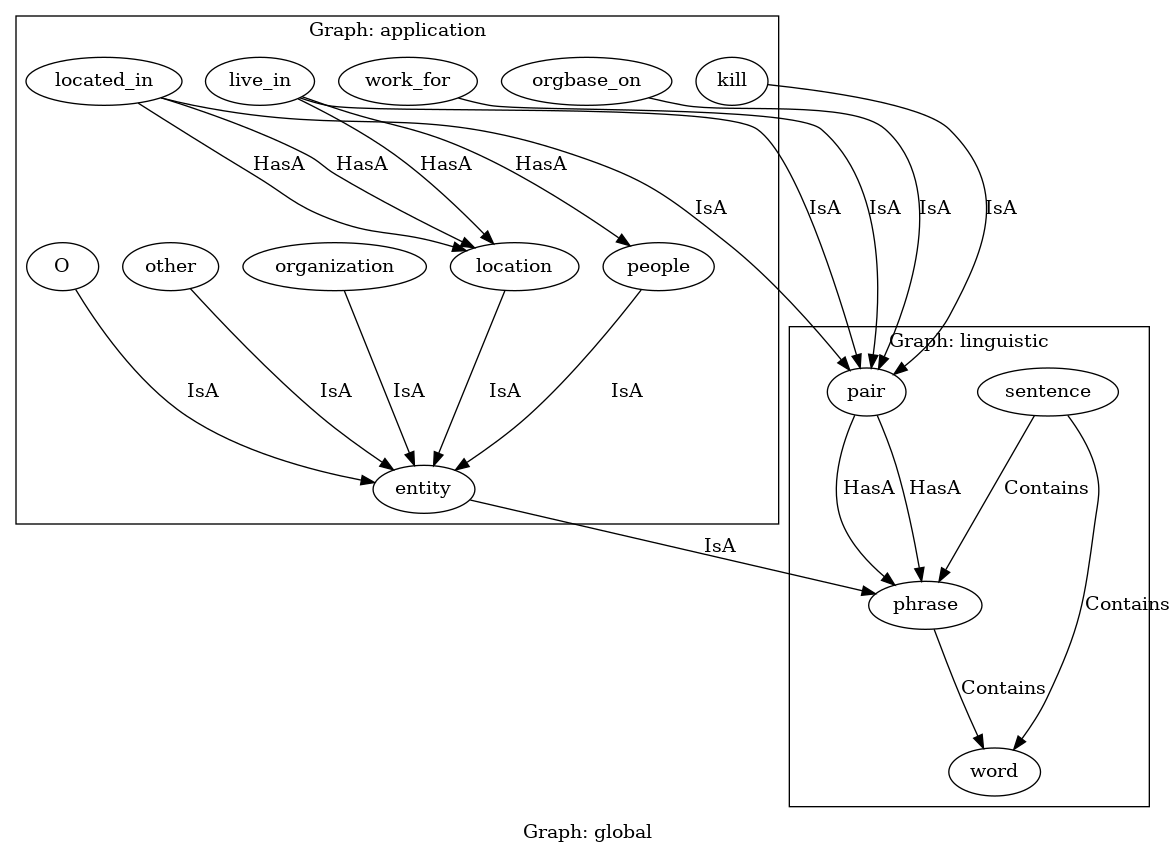}
    \caption{The domain knowledge used for the named entity and relation extraction task expressed as a graph in \framework{}}
    \label{fig:ner_graph}
\end{figure*}

\begin{figure*}[]
    \centering
    \includegraphics[width=1\textwidth]{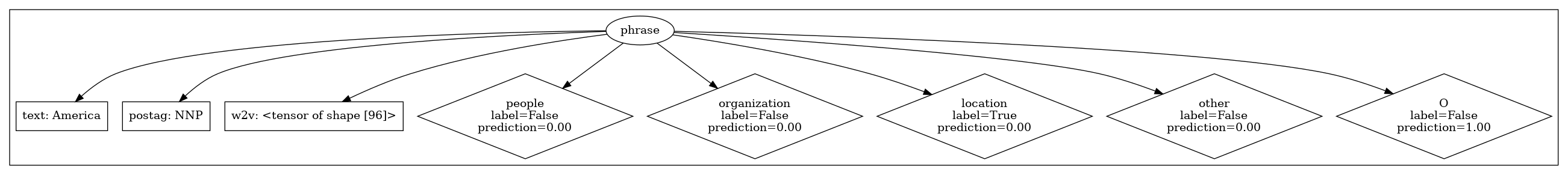}
    \caption{Sample DataNode graph populated for one single phrase from the Named Entity and Relation Extraction task.}
    \label{fig:datanode}
\end{figure*}

Figure \ref{fig:ner_graph} shows the prior structural domain knowledge expressed as a graph and used in this example. It contains the basic concepts such as `sentence`, `phrase`, `word`, and `pair` and the existing relationships between them alongside the possible output concepts such as `people` and `work\_for`. Figure \ref{fig:datanode} also represents a sampe DataNode graph populated for a single phrase alongside its properties and decisions.

Tables \ref{tab:25results} and \ref{tab:100results} summarize the results of the model applied to 25\% of the training set and the whole training set of the CONLL dataset respectively.

\begin{table*}[]
\begin{tabular}{|l|lll|lll|lll|}
                                                         & \multicolumn{3}{l|}{Precision} & \multicolumn{3}{l|}{Recall} & \multicolumn{3}{l|}{F1}    \\ \hline
                                                         & Entity   & Relation  & All    & Entity  & Relation & All   & Entity & Relation & All   \\ \hline
Baseline                                                  & 0.898    & 0.960     & 0.933  & 0.7619  & \textbf{0.807}    & 0.787 & 0.818  & \textbf{0.872}    & 0.848 \\
\begin{tabular}[c]{@{}l@{}}Baselin + \\ ILP\end{tabular} & 0.896    & \textbf{0.986 }    & \textbf{0.946}  & \textbf{0.816}   & 0.77     & \textbf{0.791} & \textbf{0.847}  & 0.86     & \textbf{0.854} \\
Baseline + IML                                           & 0.8851   & 0.979     & 0.937  & 0.746   & 0.712    & 0.727 & 0.8    & 0.82     & 0.811 \\
Baseline + PD                                            & \textbf{0.9}      & 0.952     & 0.929  & 0.765   & 0.789    & 0.779 & 0.821  & 0.859    & 0.842
\end{tabular}
\caption{The results on the 25\% of the data on Conll benchmark.}
\label{tab:25results}
\end{table*}

\begin{table*}[]
\begin{tabular}{|l|lll|lll|lll|}
                                                         & \multicolumn{3}{l|}{Precision} & \multicolumn{3}{l|}{Recall} & \multicolumn{3}{l|}{F1}    \\ \hline
                                                         & Entity   & Relation  & All    & Entity  & Relation & All   & Entity & Relation & All   \\ \hline
Baseline                                                 & 0.909    & 0.954     & 0.934  & 0.824   & 0.914    & 0.874 & 0.86   & 0.934    & 0.901 \\
\begin{tabular}[c]{@{}l@{}}Baselin + \\ ILP\end{tabular} & \textbf{0.911}    & \textbf{0.989}     & \textbf{0.954}   & \textbf{0.855}   & 0.903    & \textbf{0.882} & \textbf{0.877}  & \textbf{0.944}    & \textbf{0.914} \\
Baseline + IML                                           & 0.904    & \textbf{0.989}     & 0.951  & 0.831   & 0.884    & 0.861 & 0.86   & 0.933    & 0.901 \\
Baseline + PD                                            & 0.910    & 0.934     & 0.923  & 0.827   & \textbf{0.915}    & 0.876 & 0.862  & 0.924    & 0.897
\end{tabular}
\caption{The results on 100\% of data on Conll benchmark.}
\label{tab:100results}
\end{table*}

\subsection{Question Answering}
WIQA dataset contains 39,705 multiple choice questions regarding cause and effects in the context of a procedural paragraph. The answer is always either \textit{is less}, \textit{is more}, or \textit{no effect}.
To model this task in \framework{}, we define \textit{paragraph}, \textit{question}, \textit{symmetric}, and \textit{transitive} concepts. Each \textit{paragraph} comes with a set of \textit{question}s. As explained by \citeauthor{asai2020logicguided} enforcing constraints between different question answers can be beneficial to the models' performance. Here, two questions can be the opposite of each other with a symmetric relationship between their answers, or three questions may introduce a chain of reasoning of cause and effects leading to a transitivity property among their answers.
Following is a sample constraint defined in \framework{} to represent the symmetric property between questions.
\begin{minted}[xleftmargin=1.5em,fontsize=\small,frame=lines,linenos,breaklines,escapeinside=!!]{python}
symmetric.has_a(arg1=question, arg2=question)
ifL(is_more('x'), is_less(path=('x', arg2))
\end{minted}
\begin{table}[]
\centering
\begin{tabular}{|l|c|}
		Model  & Test Accuracy \\ \hline
		
		 Baseline & 74.22\% \\
 		 Baseline + IML & 75.49\% \\
 		 Baseline + PD & 76.59\% \\
		 Baseline + ILP & 79.05\% \\
		 Baseline + PD + ILP & \textbf{80.35\%} \\
\end{tabular}
\caption{Results of accuracy on WIQA dataset}
\label{tab:WIQA_FULL}
\end{table}

The results for the WIQA dataset are shown in table \ref{tab:WIQA_FULL}. As it is shown, IML incorporates the constraints in itself and improves the baseline a little. Primal Dual does a better job compared to IML in improving the accuracy with the help of the constraints. However, it is with the help of ILP that Baseline and Primal Dual achieve the best results.



\end{document}